\documentclass[sigconf]{acmart}

\usepackage{algorithm}
\usepackage{algpseudocode}
\usepackage{amsmath}
\usepackage{graphics}
\usepackage{epsfig}

\AtBeginDocument{%
  \providecommand\BibTeX{{%
    \normalfont B\kern-0.5em{\scshape i\kern-0.25em b}\kern-0.8em\TeX}}}

\setcopyright{acmcopyright}
\copyrightyear{2018}
\acmYear{2018}
\acmDOI{10.1145/1122445.1122456}

\acmConference[KDD '20]{KDD '20: The 26th ACM SIGKDD Conference on Knowledge Discovery and Data Mining}{Augut 22--27, 2020}{San Diego, CA}
\acmBooktitle{KDD '20: The 26th ACM SIGKDD Conference on Knowledge Discovery and Data Mining,
Augut 22--27, 2020, San Diego, CA}
\acmPrice{}
\acmISBN{}
\begin{document}
\fancyhead{}

\title{Large-Scale Training System for 100-Million Classification at Alibaba}


\author{Liuyihan Song, Pan Pan, Kang Zhao, Hao Yang, Yiming Chen, Yingya Zhang, Yinghui Xu, Rong Jin}
\affiliation{Machine Intelligence Technology Lab, Alibaba Group}
\email{liuyihan.slyh,panpan.pp,zhaokang.zk,yh136073,charles.cym,yingya.zyy,renji.xyh,jinrong.jr@alibaba-inc.com}

% \affiliation{%

% \affiliation{%

% \affiliation{%

\renewcommand{\shortauthors}{Song, et al.}
\renewcommand{\algorithmicrequire}{\textbf{Input:}} % Use Input in the format of Algorithm
\renewcommand{\algorithmicensure}{\textbf{Output:}} % Use Output in the format of Algorithm

\begin{abstract}
In the last decades, extreme classification has become an essential topic for deep learning. It has achieved great success in many areas, especially in computer vision and natural language processing (NLP). However, it is very challenging to train a deep model with millions of classes due to the memory and computation explosion in the last output layer. In this paper, we propose a large-scale training system to address these challenges. First, we build a hybrid parallel training framework to make the training process feasible. Second, we propose a novel softmax variation named KNN softmax, which reduces both the GPU memory consumption and computation costs and improves the throughput of training. Then, to eliminate the communication overhead, we propose a new overlapping pipeline and a gradient sparsification method. Furthermore, we design a fast continuous convergence strategy to reduce total training iterations by adaptively adjusting learning rate and updating model parameters. With the help of all the proposed methods, we gain 3.9$\times$ throughput of our training system and reduce almost 60\% of training iterations. The experimental results show that using an in-house 256 GPUs cluster, we could train a classifier of 100 million classes on Alibaba Retail Product Dataset in about five days while achieving a comparable accuracy with the naive softmax training process.

\end{abstract}



\maketitle

\section{Introduction}
In recent years extreme classification has attracted significant interests in the areas of computer vision and NLP. It introduces a vanilla multi-class classification problem where the number of classes is significantly large. Such a large classifier has achieved remarkable successes, especially in applications like face recognition \cite{sun2014deep} and language modeling \cite{chen2016strategies}, when training on the industry-level datasets.

At Alibaba, the Retail Product Dataset contains up to billions of images across 100 million classes. Each image is labeled at stock keeping unit (SKU) level. We want to build a 100 million-level extreme classification system with the dataset to improve the recognition abilities of our vision system.

% \begin{figure*}[h]
\begin{figure}
  \centering
  \includegraphics[width=\linewidth]{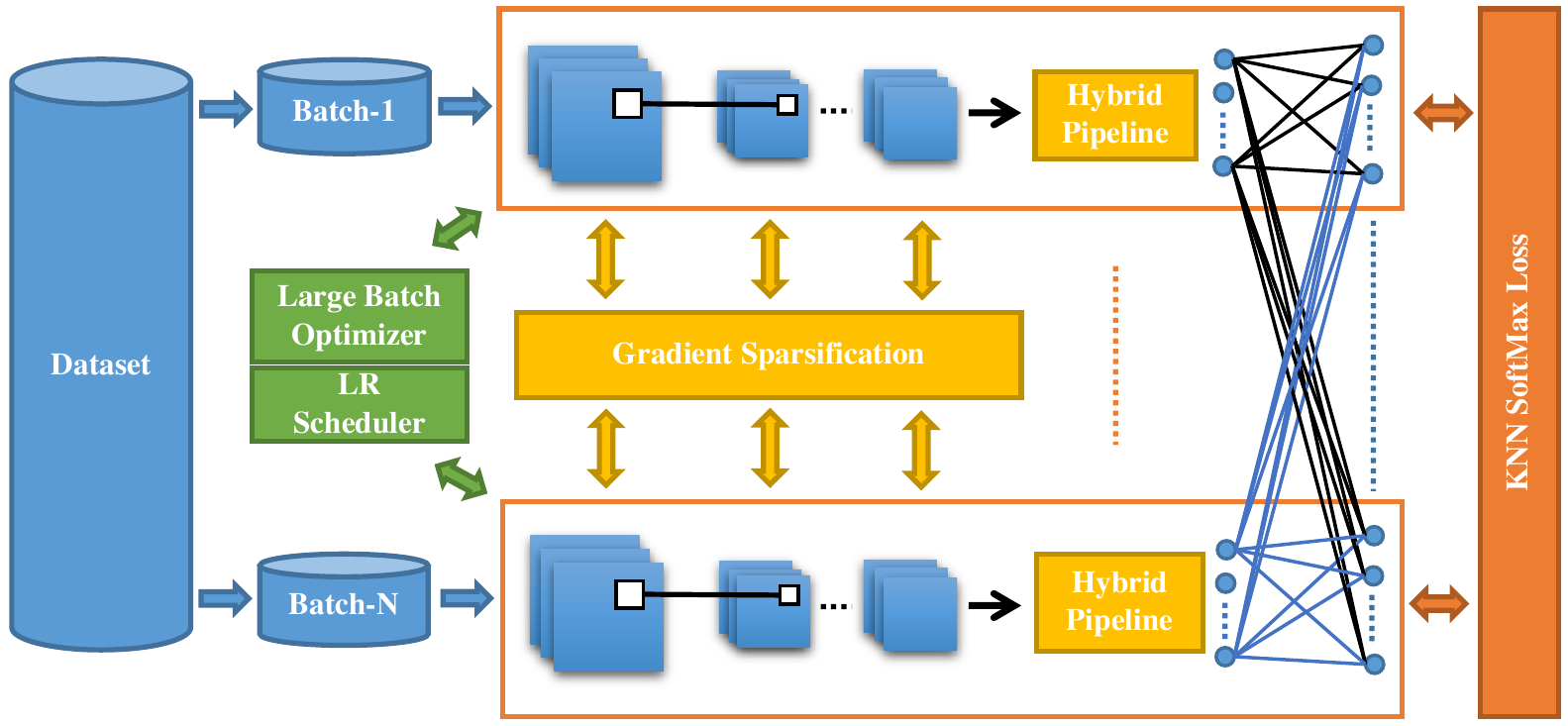}
  \caption{Overview of the overall extreme classification system architecture. It contains three major components: (i) KNN softmax loss module for fast computation and derivation. (ii) Communication module including hybrid pipeline and gradient sparsification. (iii) Fast convergence module with a large batch optimizer and a learning rate scheduler.}
  \label{fig:overview}
\end{figure}
% \end{figure*}

However, building an extreme classification system poses a number of challenges as follows:

{\textbf{Memory and computation costs:}} As the parameter size of the last fully connected layer is proportional to the number of classes, it may go beyond the GPU memory capacity when training with a large classifier straightforwardly. Also, the computational cost will be significantly increased, which is approximately measured by the dot products between the class weights and input features.

{\textbf{Communication overhead:}} To accelerate the training process as much as possible, one could add more GPU machines to process more data samples synchronously. However, as the number of GPU machines increases larger, the overhead of communication among machines will become the bottleneck of training speed.

{\textbf{Convergence:}} In parallel training, synchronous stochastic gradient descent (SGD) is often used in training the large-scale deep neural networks. With an increase in the number of GPU nodes, training with large batch usually results in lower model accuracy. Moreover, the convergence speed is unacceptable with a large number of epochs, e.g., 90 epochs in ImageNet-1K training \cite{he2016deep}.

Prior efforts tackle these difficulties in several ways. To reduce the resource cost, feature embedding methods \cite{bhatia2015sparse, tagami2017annexml, nigam2019semantic} are proposed. These methods project the inputs and the classes into a small dimensional subspace instead of using a large fully connected layer at last. Nevertheless, training embedding models need a pairwise loss function, which uses a large number of training pairs and needs carefully designed negative sampling. Another solution is hierarchical softmax \cite{goodman2001classes, mikolov2013efficient}, but it is often difficult to extend these methods along to other domains and cannot guarantee accuracy on image classification tasks. Meanwhile, hierarchical softmax is not parallel friendly, which means it is hard to support multi-GPU training. To our best knowledge, using a standard softmax with cross-entropy based classifier could solve these issues.

In this paper, considering the drawbacks of those methods mentioned above, we propose an extreme classification system by collaboration of algorithm and engineer teams at Alibaba. Unlike \cite{jia2018highly, you2018imagenet} using a data parallel framework to train ImageNet-1K in a few minutes, we use a hybrid parallel framework to alleviate model partition in a GPU cluster. In this way, we can train such a ``big head'' neural network using a standard softmax with cross-entropy loss. Figure \ref{fig:overview} shows the overview of our extreme classification system architecture. We conclude our contributions as follows,

1) We introduce an effective softmax implementation named KNN softmax to classify 100 million classes of images straightforwardly. Compared with the selective softmax \cite{zhang2018accelerated} or MACH \cite{medini2019extreme}, our approach achieves the same accuracy as standard softmax. Furthermore, our proposed method saves computation and GPU memory, which improves training speed correspondingly.

2) We propose a new communication strategy, which includes an overlapping pipeline and a gradient sparsification method. For our hybrid parallel training framework, this communication strategy reduces the overhead and accelerates training speed.

3) As large batch training plays an essential part in our training framework, we propose a new training strategy to update model parameter and adjust learning rate adaptively. In this way, we could significantly reduce our training iterations and achieve a comparable accuracy with the naive softmax training process.

The rest of the paper is organized as follows. Section 2 briefly reviews the related work. The proposed framework and methods are detailedly described in Section 3. Experimental evaluations are shown in Section 4. Finally, Section 5 concludes this paper.

\section{Related work}
Building an extreme classification system includes four primary sections. Firstly, we need to develop a parallel training method for extreme classification. Secondly, an accuracy-lossless softmax computation algorithm should be carefully designed. Thirdly is to build an efficient communication strategy in a large GPU cluster. Besides, fast convergence is also essential for an efficient training process. Taking the applied techniques into account, prior practices of building an extreme classification system can be concluded in the following aspects.

% \subsection{Parallel Training}
\textbf{Parallel Training:} Recently, \cite{krizhevsky2014one} proposes a training scheme which uses data parallelism and model parallelism together to parallelize the training of convolutional neural networks with stochastic gradient descent (SGD). Deng et al. \cite{deng2019arcface} employ a parallel training strategy to support millions-level identities on a single machine efficiently. In our scenarios, we extend the hybrid parallel training scheme to a larger GPU cluster. Meanwhile, we also optimize the training pipeline to accelerate training.

% \subsection{Selective Softmax}
\textbf{Softmax Variations:} 1) Selective Softmax: \cite{zhang2018accelerated} proposes a new method to solve the extreme classification problem. In particular, they develop an effective method based on the dynamic class level to approach the optimal selection. This method has two drawbacks. Firstly, the method is not a completely GPU implementation since the entire $\textbf{W}$ is maintained in CPU RAM. Moreover, the performance of selective softmax is inferior to the full softmax, especially in large-scale experiments, which is not acceptable in practice. 2) Merged-Average Classifiers via Hashing: To solve the $k$-class classification problem, a simple hashing based divide-and-conquer algorithm, MACH (Merged-Average Classification via Hashing) \cite{medini2019extreme}, is proposed. Compared with the traditional linear classifier, it only needs a small model size. However, the method is still unable to get a comparable performance compared with standard softmax. As stated in \cite{medini2019extreme}, in ImageNet dataset, MACH achieves an accuracy of 11\%, while full softmax achieves the best result of 17\%.

% \subsection{Efficient Communication}
\textbf{Efficient Communication:} Large-scale distributed parallel SGD training \cite{bottou2018optimization} requires gradient/parameter synchronization among tens of nodes. With increasing numbers of nodes, communication overhead becomes the bottleneck and prohibits training scalability. As centralized network frameworks like parameter server \cite{li2014scaling} are limited by network congestion among central nodes, decentralized network frameworks with collective communication operations (all-reduce, all-gather, etc.) are widely used in large-scale distributed training. Besides utilizing expensive high-performance networks (100 Gbps Ethernet, InfiniBand, etc.), multiple methods have been proposed to mitigate communication overhead. Pipelining overlaps bottom layers gradient computation and top layers gradient communication during backpropagation. It has been widely used in distributed machine learning frameworks such as PyTorch \cite{paszke2019pytorch} and MXNet \cite{chen2015mxnet}. Recently, gradient compression methods that reducing transmitted bits per iteration draw much attention. Sparsification \cite{aji2017sparse, lin2017deep} methods selected part of gradients based on the magnitude and conserved ImageNet-1K accuracy with gradient sparsity up to 99.9\%. Quantification \cite{karimireddy2019error} methods encoded gradients into 1-bit, thus achieving up to 1/32 compression ratio. Low-rank factorization \cite{vogels2019powersgd} communicated a low-rank lossy approximation of gradients to reduce network traffic.

% \subsection{Fast Convergence}
\textbf{Fast Convergence:} Early works mostly focus on the learning rate adjustment to deal with large batch training. \cite{goyal2017accurate} set the initial learning rate as a function of batch size according to a linear scale-up rule. The method managed to apply the approach to train a ResNet-50 network on ImageNet-1K with a batch size of 8,000 over 256 GPUs. In \cite{you2018imagenet}, training with a much larger batch size of 32K can be finished in 20 minutes with LARS. Since the gradients of DNN network in early steps may vary significantly, large learning rate may cause divergence. To avoid divergence, a warm-up strategy that increases the learning rate gradually from a very small value is proposed \cite{goyal2017accurate}. However, all the above techniques are only proved to work in the training of ResNet-50 on ImageNet-1K, and less attention is paid to the training of larger or more complex models on other datasets.

\section{100 Million Classification}
\subsection{System Architecture}
To train a classifier of 100 million classes, how to store the large fully connected (fc) layer is the first and primary problem to be addressed. Assuming the dimension of input feature is 512, the total GPU memory cost of the fully connected layer is about 190 GB which cannot be fed into a single GPU. Therefore, training such a large classifier is barely impossible using a data parallel training framework.

As mentioned in \cite{krizhevsky2014one}, we split the large fc layer into different sublayers and place each sublayer into different GPUs in a model parallel way. It has two advantages: 1) The computation cost at the fully connected layer can be reduced proportionally to the number of GPU used. 2) The communication overhead of synchronizing gradient can be reduced significantly since the fully connected layer is updated locally compared to the data parallel training.

\begin{figure}
  \centering
  \includegraphics[width=\linewidth]{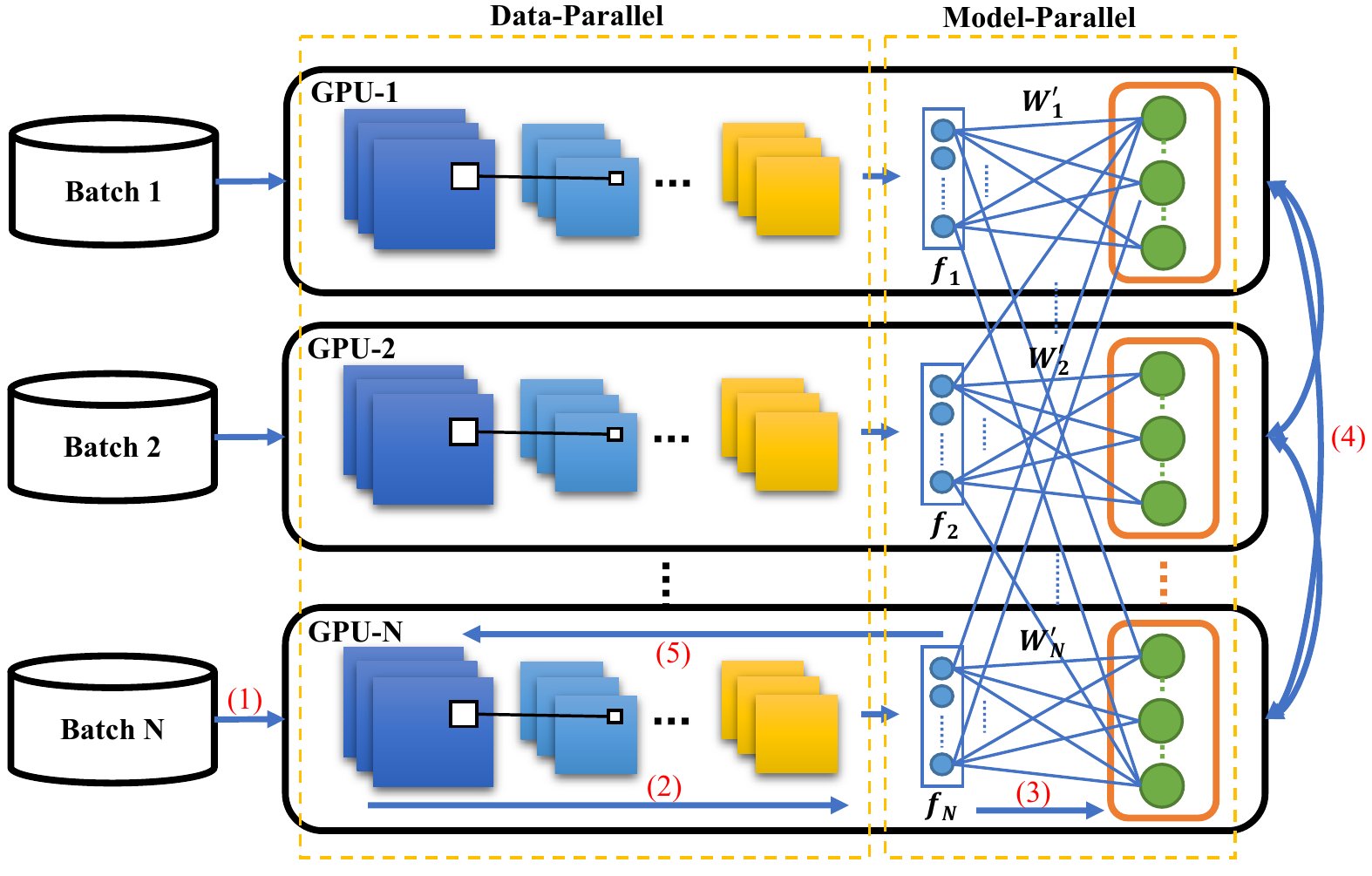}
  \caption{Hybrid parallel training framework}
  \label{fig:hyb}
\end{figure}

Since the fully connected layer is split in model parallel way, we can reuse the remaining memory space of each GPU for the feature extraction part before the fully connected layer. Meanwhile, the feature extraction part is trained in data parallelism. Therefore, the proposed framework of this work belongs to hybrid parallel training. Figure \ref{fig:hyb} presents the overall hybrid parallel training framework.

As depicted in Figure \ref{fig:hyb}, we use GPU-$N$ to elaborate on how the hybrid parallel training works: 1) Data batch-$N$ is fed into GPU-$N$. 2) GPU-$N$ uses convolutional neural networks to extract the features of batch-$N$, and then gathers the features extracted from other GPUs. 3) GPU-$N$ forwards features through the $N$-th sublayer of fully connected layer 4) Distributedly compute softmax with cross-entropy loss using all GPUs. 5) GPU-$N$ backpropagates gradients through the whole network. It is worth mentioning that the gradient of $f_N$ needs to merge with the corresponding item from other GPUs. 6) For model parallel parts, GPU-$N$ updates weights by local gradient; For data parallel parts, it merges the corresponding gradient of feature extraction part all over the GPUs then update weights (step-6 did not show in Figure \ref{fig:hyb}).

Additionally, as mixed-precision training \cite{micikevicius2017mixed} has been widely used in computer vision and NLP tasks without sacrificing accuracy, we adopt this method to accelerate our training. For our hybrid parallel training framework, we convert all the layers except batchnorm \cite{ioffe2015batch} to float16, and gradients are calculated by float16 too. Meanwhile, all parameters have a copy in float32 for parameter updating. Besides, we use loss scaling \cite{micikevicius2017mixed} to preserve small gradient values during training.

By implementing a hybrid parallel training framework on 100 million classification, we run an end-to-end training profiling on the in-house GPU cluster. According to the profiling, almost 80\% of the GPU memory is consumed by the fully connected layer, which leads to a low training throughput since less memory can be used for feature extraction parts. Therefore, new methods that reduce memory and computation costs is strongly needed in training the large classifier. The following sections will describe the proposed methods that overcome the difficulties of such a training task.

\subsection{KNN Softmax}
As mentioned above, the last fully connected layer consumes large amounts of GPU memory. By conducting experiments of training a classifier of 100 million classes, it is also noted that in each iteration more than 80\% of the time is spent in the softmax stage (mainly including fc forward, softmax forward, softmax backward and fc backward), and over 10 GB of GPU memory is used for the output space of the last fc layer.

In order to further improve the throughput of the system, we propose a new method called {\it KNN softmax}. Specifically, we adopt the active classes to speed up the softmax stage and save the memory demand as in \cite{zhang2018accelerated}. What's more, through combining normalization strategy and a KNN graph-based selection approach, we achieve the lossless performance compared with the standard softmax, which is essential in practice. Finally, we provide a completely GPU-based training pipeline for our method.

\subsubsection{Active classes selection}

Inspired by selective softmax \cite{zhang2018accelerated}, we also select the active classes for each mini-batch and calculate forward/backward based on them. Differently, we adopt a new way to do the active classes selection. Let N be the total number of classes, we denote $\textbf{W} \in \textrm{R}^{N \times D}$ as the weight parameters of last fc layer, where each row $\textbf{w}_j$ represents the weight vector of the $j$-th class. For each training example $\textbf{x}^{i}$, we use its weight vector $\textbf{w}_{y^{i}}$ ($y^{i}$ is the label of $\textbf{x}^{i}$), instead of $\textbf{x}$ itself, to select active classes. On this condition, we can quickly fetch the active classes from a $k$-nearest neighbor (KNN) graph of $\textbf{W}$ that we build in advance to avoid the time complexity of searching the active classes using $\textbf{x}$. 

Specifically, we compute the $L_2$ normalization of $\textbf{X}$ (the extracted feature of mini-batch) and $\textbf{W}$ in the training process first. Next, a KNN graph for the $\textbf{W}_{norm}$ (the $L_2$ normalization of $\textbf{W}$) is constructed (the building process will be described in the following section). Assuming we already built this graph, in each iteration of training process, we can quickly get the active classes of mini-batch: $\textbf{W}_{active}=[list_{y^{1}},...,list_{y^{m}}]$ where $list_{y^{i}}$ is the KNN result of $\textbf{w}_{y^{i}}$. Considering $\textbf{W}$ has been normalized, $\textbf{w}_{y^{i}}$ must be ranked first in the $list_{y^{i}}$. After that, we will remove the duplicated $\textbf{w}$ from $\textbf{W}_{active}$, then compare the number of the remaining active classes with $M$ (the number setting of active classes for each iteration) to select the final active classes. Algorithm \ref{alg:acs} summarizes the KNN Graph-based Active Classes Selection.

\begin{algorithm} %[t]
\caption{KNN Graph-based Active Classes Selection} 
\label{alg:acs} 
\begin{algorithmic}[1] 
\Require 
A KNN graph, $G = [list_0,...,list_{N-1}] \in \textrm{R}^{N \times K}$;
The entire weight vector $\textbf{W} \in \textrm{R}^{N \times D}$; 
The mini-batch feature $\textbf{X}_{norm} \in \textrm{R}^{m \times D}$; 
The number of selected active classes M;
\Ensure 
The active classes of the current mini-batch, $W_{active}^*$; 
\State Initialize active classes set $W_{active} = \varnothing$; 
\label{code:fram:extract} 
\For {each sample $\textbf{x}^i$ in $\textbf{X}_{norm}$}
\State insert $list_{y^i}$ into $W_{active}$
\EndFor
\State $W_{active}^{'} = duplicate(W_{active})$; 
\label{code:fram:add} 
\If {$W_{active}^{'}.size$ $<$ M}
\State $W_{random}$ = random sample ($M - W_{active}^{'}.size$) weight from $\overline {W}_{active}^{'}$ (the non-choosen weight in $\textbf{W}$)
\State $W_{active}^*$ = $W_{active}^{'} + W_{random}$  
\Else
\State $W_{active}^*$ = get M weight from $W_{active}^{'}$ based on their ranking score.
\EndIf
\State \Return $W_{active}^*$; 
\end{algorithmic} 
\end{algorithm}

\subsubsection{Distributed graph building}

Generally, one like to use approximate nearest neighbor (ANN) methods to build graph \cite{zhao2019large}, which could achieve the tradeoff between performance and time consuming. However, we empirically find the quality of KNN graph has a great influence on the final accuracy. The ANN graph can not guarantee all the nearest neighbors are recalled. Once some nearest neighbors are lost, it will inevitably bring about the loss of the active classes of certain samples in the training process, which will lead to the difference of final performance compared to the full softmax. Consequently, we utilize linear search to ensure the precision of nearest neighbors. 

The brute-force graph building is a very time-consuming process, so we will rebuild the graph after a long iterations. Moreover, in order to save the computational resources, we will reuse the GPU of training to construct graph (the training will be suspended at that time). 

\begin{figure} %[b]
  \centering
  \includegraphics[width=\linewidth]{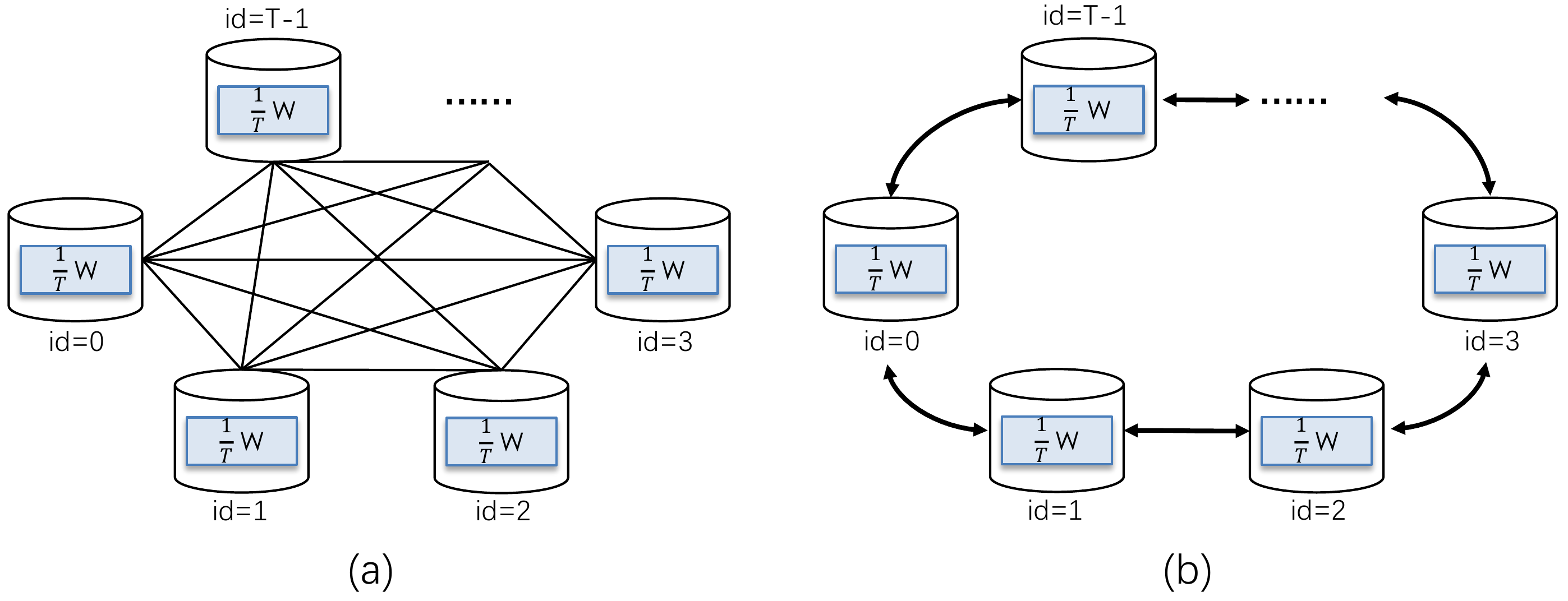}
  \caption{Distributed graph building with multiple GPUs.}
  \label{fig:dgb}
\end{figure}

With the normalization of $\textbf{W}$, the Euclidean distance and inner product are equivalent, and the inner product calculation is a matrix multiplication on CUDA , which is easy to be implemented. As mentioned above, $\textbf{W}$ is stored in different nodes, so we use the ring structure in Figure \ref{fig:dgb} (b) to transfer local $\textbf{w}$ between different nodes. After the node gets the local $\textbf{w}$ from the former one, it will calculate the mm operation and update its NN list. Then the received local weight will be sent to the next node. Compared with gathering all $\textbf{w}$ into one node (Figure \ref{fig:dgb} (a)), our method can avoid the burst of GPU memory due to too large $\textbf{W}$ (matrix multiplication also takes up a lot of temporary memory).

In addition, we transform the $\textbf{W}$ from float32 to float16, and use the TensorCore\footnote{\url{https://www.nvidia.com/en-us/data-center/tensorcore/}} to accelerate the matrix multiplication. For the sake of the graph quality, we will recall $k'$-nearest neighbor ($k'$ is larger than $k$), and then perform the standard float32 calculation based on the $k'$ neighbors to get the final kNN (the float32 calculation here can almost be ignored). TensorCore can speed up the whole process about three times.

In practice, we rebuild the graph after one epoch training is finished (To make it fair, we will take the graph building time into account when evaluating the efficiency of KNN softmax in the experiment section). Thanks to the efficient pipeline of GPU implementation, the time consuming of 100 million graph building can be controlled within 0.75 hours with 256 V100.

\subsubsection{Efficient implementation}

After the graph construction, we intend to make the training completely on the GPU as well. Considering the classification of 100M, we set $K = 1000$ to ensure the performance, and then the graph size is $100 M \times 1000$, which is about 372 GB. When training begins, each node needs to query the complete graph to select the active classes. In other words, the complete graph should be stored in every node, which is very hard to be loaded into CPU memory, let alone the GPU memory.

So as to make full use of GPU to train our classifier, we present two steps to solve the graph storage problem (the GPU we used is 32 GB V100):
\\\textbf{(i) Graph compression.} On each node, keeping a complete graph is always redundant. Because the $\textbf{w}$ that are not on this node cannot be selected by any mini-batch gathering on the node. As a result, we can delete all the redundant $\textbf{w}$ index from the graph stored on one node. Suppose we use 256 GPU for training, the storage can be compressed to 372 GB / 256 = 1.45 GB on average;
\\\textbf{(ii) Quick Access.} With the graph compression, the neighbors num (K) of each $\textbf{w}$ in the graph is no longer the same. So we turn a two-dimensional tensor ($100M \times K$) into a one-dimensional tensor. A new problem arises: we can not get $\textbf{W}_{active}$ efficiently as before since the one-dimensional tensor cannot be accessed directly with the label as index. To tackle this problem, we added a new kernel function into PyTorch framework to quickly access the compressed graph. More concretely, we first store the new K value of each $\textbf{w}$ into a tensor, and use another tensor of the same size to accumulate the K value (the accumulation result is the offset of the $\textbf{w}$ in the compressed graph). In the training process, we use different threads to find the offset of each sample in the compressed graph. \\

We summarize the core \textbf{GPU Pipeline} of KNN softmax as follows, which mainly consist of three steps:
\begin{enumerate}
\item \textbf{Graph Building:} We use distributed GPUs to compose the graph of $\textbf{W}_{norm}$, and adopt method (i) \emph{graph compression} to compress and store the graph on all GPUs.\\
\item \textbf{Normalization:} Normalization of $\textbf{X}$ and $\textbf{W}$ in GPU is executed during training process.\\
\item \textbf{Active Classes Selection:} For the normalized mini batch $\textbf{X}_{norm}$, method (ii) \emph{quick access} is employed to implement active classes selection on GPU.\\
\end{enumerate} 

Though we both select active classes, our KNN softmax is totally different from selective softmax \cite{zhang2018accelerated} in the following three aspects. 1) We use KNN graph to do the active classes selection, instead of the Hashing Forest used by selective softmax; 2) Selective softmax is not completely GPU implemented; 3) Last but not least, we maintain the same precision as the full softmax, which is hard to be achieved by selective softmax.

\subsection{Communication Strategy}
As KNN softmax significantly reduces GPU memory and computation cost, communication overhead becomes the bottleneck in large-scale distributed parallel SGD training. Based on the training profiling of our hybrid parallelism framework, we applied an efficient hybrid parallel pipelining to introduce more overlapping in the forward and backward stages. Besides, we implemented an efficient gradient sparsification method \cite{lin2017deep} to reduce transmitted bits during backpropagation. Under the premise of ensuring model convergence, our strategies reduce wall clock time per iteration and improve the throughput of large-scale mini-batch training.

\subsubsection{Hybrid parallel pipelining}
\begin{figure} %[h]
  \centering
  \includegraphics[width=\linewidth]{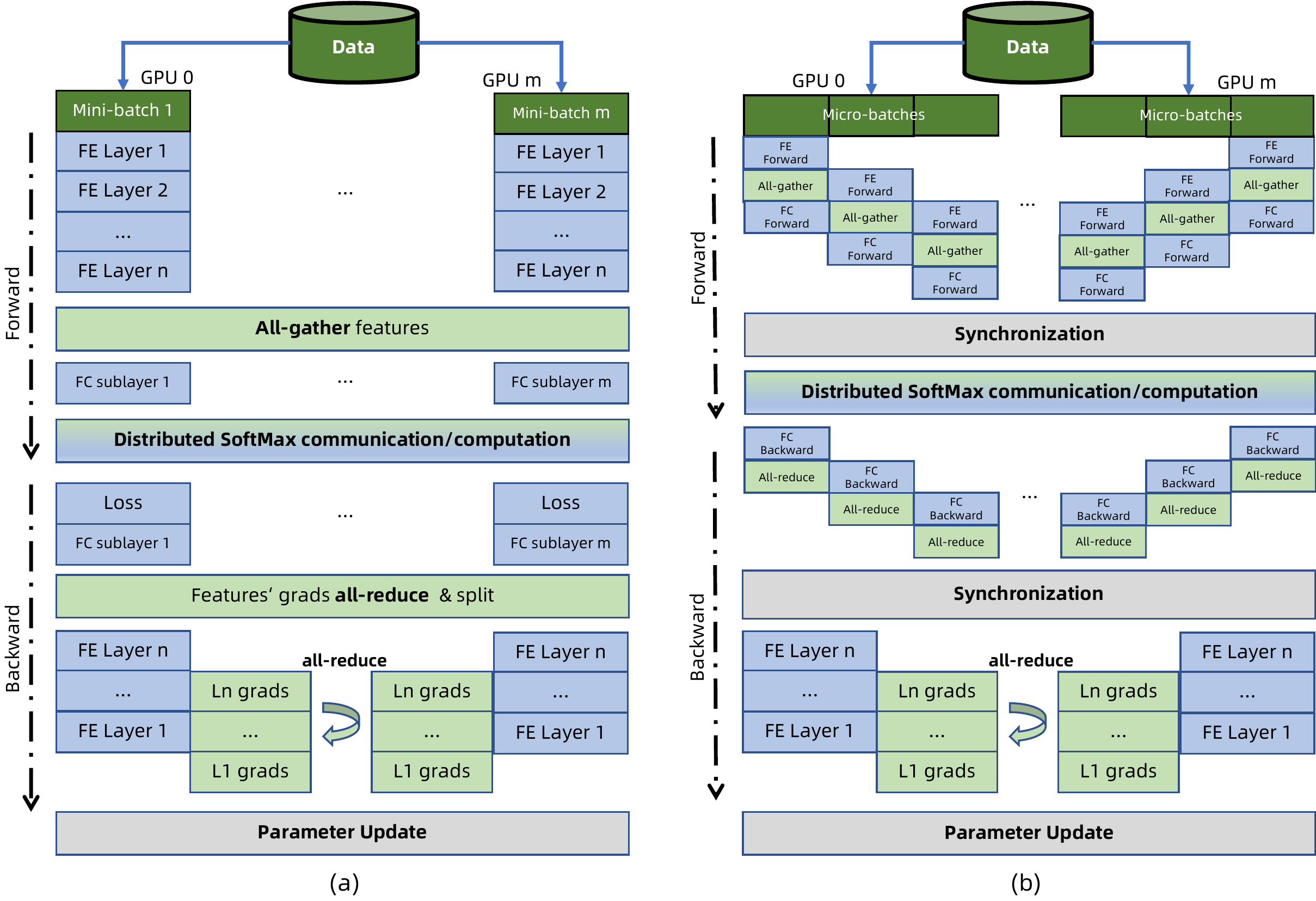}
  \caption{(a) is the baseline communication/computation ovarlapping for hybrid parallel training and (b) is our proposed hybrid parallel overlapping pipeline. (The blue chunk indicates computation while the green one indicates communication.)}
  \label{fig:pipeline}
\end{figure}

Data parallelism only involves inter-node communication to synchronize gradients. Typical pipelining overlaps the synchronization and computation of gradients during backpropagation. While under our large-scale classification hybrid parallel framework, communication involves a) the transmission of features from the data parallel feature extraction (FE) part to fc layer; b) communicate among fc layer to compute softmax; c) gradient synchronization during backpropagation.  b) is insignificant due to its tiny message size. As shown in Figure \ref{fig:pipeline} (a), the fc sublayers are idle until all the feature extraction parts compute features and accumulate through all-gather communication and vice versa during backpropagation. 

To overlap computation and communication in our hybrid parallel framework, we divide the mini-batch samples into micro-batch samples with asynchronous computation and communication among micro-batch samples. Figure \ref{fig:pipeline} (b) shows our pipelining strategy. The fc layers collect the features among different nodes with an all-gather communication once a micro-batch forward computation completes, thus overlapping forward computation of the feature extraction part. In the backpropagation, we overlap the fc layer gradient computation and all-reduce communication among micro-batches. For the feature extraction part, we follow the common data parallelism pipelining method. With our hybrid parallel overlapping pipeline, we can achieve more overlapping between communication and computation. Furthermore, the divided micro-batches save GPU memory usage and enable a larger batch size in a single GPU.

\subsubsection{Efficient layer-wise top-$k$ sparsification}
Hybrid parallel pipelining overlaps feature extraction net and fc sublayers with the micro-batches split. For the feature extraction net backpropagation, gradient compression methods can be utilized to further mitigate communication overhead. Deep gradient compression \cite{lin2017deep} (DGC) performs a layer-wise top-$k$ selection among layers' tensor and only communicate selected gradients among nodes. With tricks including momentum correction, momentum factor masking, DGC raises the communication tensor sparsity to 99.9\% while preserving the accuracy of training ResNet-50 model on the ImageNet-1K dataset. DGC is not widely used in the industry, mainly because the layer-wise top-$k$ selection is time-consuming. For example, \cite{sun2019optimizing} implemented DGC in 56 Gbps Ethernet and showed no throughput improvement. DGC proposed a sampling top-$k$ to reduce selection time. However, it introduces approximation of top-$k$ selection and is still inefficient given a low communication-to-computation ratio model like ResNet. 

\begin{figure} %[h]
  \centering
  \includegraphics[width=\linewidth]{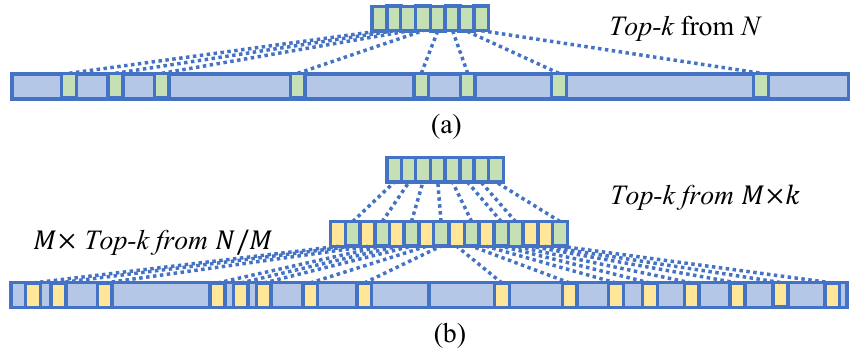}
  \caption{Divide and conquer top-$k$ selection}
  \label{fig:topk}
\end{figure}

To deal with the top-$k$ selection computation overhead, we apply a divide-and-conquer top-$k$ selection and grouping tensors with similar size, which makes full use of GPU parallel computing ability and greatly reduces the computation overhead without any approximation. As shown in Figure \ref{fig:topk}, we divide single top-$k$ selection from large tensor into two steps. First, we split a large tensor into M small chunks and select top-$k$ from every chunk ($M \times K$) simultaneously. Then, a second top-$k$ selection is carried out from selected $M \times K$ tensor. Grouping tensors with similar size makes the layer-wise top-$k$ selection more highly integrated. With our top-$k$ selection implementation, the extra computation overhead can be negligible. Combined with hybrid parallel pipelining, the end-to-end iteration wall clock time is further reduced.

\subsection{Fast continuous convergence algorithm}
The extremely large fc layer involves the update of tens of billions of parameters in each iteration. As a result of the limitation of GPU resources, accelerating convergence becomes challenging: 1) In order to prevent such a large model from under-fitting or over-fitting, the amount of training data will be large, resulting in a very long time of training. 2) With limited computation resources, any sophisticated learning strategies that involve trial and error are not desired.

% However, the existing training methods cannot solve the above challenges. A typical learning strategy is the piece-wise decay, which sets the learning rate to a large initial value at the beginning, and then multiplies the learning rate by a reduction factor every few iterations. In such way that we need to do lots of experiments to adjust the hyper-parameters in order to achieve the best accuracy. Although another adaptive optimizer, Adam \cite{kingma2014adam}, reduces the difficulty of adjusting parameters, it may bring a loss of convergence accuracy.

To address the above chanllenges, we develop a very aggressive convergence algorithm to solve the problem of fast convergence of the large-scale classification models on large-scale datasets, called fast continuous convergence strategy (FCCS). We divide the convergence strategy into global policy and local policy. The local policy takes advantage of the local learning rate calculation in LARS \cite{you2017scaling}, which enables large batch training, to overcome the inefficiency of massive data training. And the global policy aims to control the speed of model convergence, which give us the opportunity to complete the training quickly within a finite number of iterations.

The global policy mainly focuses on the adjustment of batch size and learning rate. We divide the global policy into two different phases. The first is the phase of warm-up, during which only the learning rate is adjusted while the batch size remains unchanged. The second phase includes a progressive continuous increase of batch size while we keep the learning rate at a constant value.

Let ${B}_{t}$ be the batch size of the $t$-th iteration, and $\mathbf{\eta}_{t}$ be the learning rate of the $t$-th iteration. The learning rate adjustment strategy includes a warm-up stage, that is, starting from a small learning rate during training, and gradually increasing to a large value $\mathbf{\eta}_{0}$. Then the learning rate remains constant after the warm up stage:

$$
\eta_t= \left\{
\begin{array}{lr}
\frac {t} {T_{warm}} \eta_0  & \textsf{if}\ t < T_{warm} \\
\eta_0 & \textsf{if}\ t \ge T_{warm} \\
\end{array} \right.
$$

Where ${T}_{warm}$ is the total number of iterations of warm-up.

The batch size adjustment is divided into an initialization stage and an aggressive continuous increase stage. It can be described as follows.

$$
B_t= \left\{
\begin{array}{lr}
B_0  & \textsf{if}\ t < T_{ini} \\
\lfloor f(t) \rfloor & \textsf{if}\ t \ge T_{ini} \\
\end{array} \right.
$$

Where ${f}_{t}$ is defined as follows:

$$
f(t) = B_{min}^{1} + \frac 1 2 (B_{max}^{1} - B_{min}^{1})(1 + cos(\frac {t - T_{ini}} {T_{final} - T_{ini}} \pi))
$$

During the initialization stage, the batch size keeps a small constant value ${B}_{0}$, which guarantees a sufficient update frequency in the warm-up period. During the stage of aggressive continuous increase, the batch size increases quickly as the number of iterations increases.

According to the theory of \cite{smith2017don}, somehow increasing the batch size is equivalent to reducing the learning rate, so we replace the traditional learning rate decay process by increasing the batch size. At the same time, it should be emphasized that in our method, the batch size is increased in a continuous manner. This is because if it is a discontinuous manner such as piece-wise police, more experiments are needed to be tried to clearly determine the hyper-parameter settings. Compared with these methods, our continuous growth policy avoids the choice of these hyper-parameters and only needs to control the speed of batch size growth.

Besides, to overcome the limitation of GPU memory, we apply the gradients accumulation technique to enlarge the batch size. By accumulating the gradients $n$ times without updating the parameters, the actual batch size can be considered as $n \times b$. This also brings another benefit that gets the total communication cost decreased at most to $1 / n$ of that with constant batch size.

% \begin{figure}[h]
%   \centering
%   \includegraphics[width=\linewidth]{figure/grads_accu}
%   \caption{increase the batch size by gradients accumulation. In the normal procedure, parameters are updated every time after backward. In gradients accumulation, the update is conducted only once after n iterations.}
%   \label{grads_accu}
% \end{figure}

\section{Experiments}
In this section, we conduct extensive experiments to evaluate the performance of each algorithm in our system. To conduct the evaluation for each component in our system, we randomly sample three subsets of images containing 1 million, 10 million, and 100 million classes from Alibaba Retail Product Dataset. Correspondingly We call these three datasets SKU-1M, SKU-10M, and SKU-100M for convenience. Overall information related to each dataset is listed in Table \ref{tab:datasets}.

\begin{table}[h]\fontsize{9.5pt}{\baselineskip}\selectfont
\centering
\caption{\label{tab:datasets}Overview of three datasets.}
\begin{tabular}{|l||c|c|c|} \hline
\#datasets & total classes & train samples & test samples \\ \hline
SKU-1M  	 & 1,020,250     & 51,901,530    & 9,375,825    \\ \hline
SKU-10M    & 9,890,866     & 404,974,133   & 75,657,866   \\ \hline
SKU-100M   & 100,001,020   & 2,708,042,900 & 301,754,332  \\
\hline
\end{tabular}
\end{table}

For the three datasets, we use ResNet-50 \cite{he2016deep} as the base model of the feature extraction part in our hybrid parallel training framework. The final dimensions of the features are 512. Moreover, we apply the mixed-precision training method to our training framework as default.

All of the experiments are running on the in-house GPU cluster. This cluster contains 32 machines, and each machine uses 8 NVIDIA Tesla V100 (32GB) GPUs, which are interconnected with NVIDIA NVLink. For network connectivity, the machines use a 25Gbit Ethernet network card for communication. We use PyTorch \cite{paszke2019pytorch} for our distributed training implementation.

\subsection{Evaluation of KNN Softmax}

We assess the classification accuracy and throughputs respectively. For the softmax accuracy, we compare our approach with the following state-of-the-art methods: 

\begin{itemize}
\item \textbf{Selective Softmax} \cite{zhang2018accelerated}: We use the HF-A version with $L=50$, $T=1000$ and $\tau_{cp} = 0.9$;
\item \textbf{MACH} \cite{medini2019extreme}: We set different $B$ and $R$ for different scale datasets: $B=1024, R=32$ for 1M; $B=4096, R=32$ for 10M; $B=10240, R=64$ for 100M.
\item \textbf{Full Softmax}: The traditional softmax with the hybrid parallel training framework.
\item \textbf{KNN Softmax}: This is our method proposed in this paper. We vary the $k$ (12 for 1M, 120 for 10M, 1200 for 100M) and choose the $10\%$ active classes.
\end{itemize}

To make it fair, we only compare the methods that have the ``same'' accuracy with the full softmax for the throughputs evaluation (``low'' accuracy methods are ignored since one can sacrifice accuracy to improve throughputs).

\begin{table}[h]\fontsize{9.5pt}{\baselineskip}\selectfont
\centering
\caption{\label{tab:softmaxpre}The classification accuracy of different methods in the three datasets.}
\begin{tabular}{|l||c|c|c|} \hline
\#methods &{\bf 1M} & {\bf 10M} & {\bf 100M} \\ \hline
Selective Softmax \cite{zhang2018accelerated} 	& 86.39\% & 79.02\%  & 71.98\%  \\
MACH \cite{medini2019extreme}  	& 80.11\% & 71.34\% & 59.82\%  \\
KNN Softmax   	& 87.46\%  & 80.99\%  & 74.54\%  \\ \hline
Full Softmax   	& 87.43\%  & 81.01\%  & 74.52\%   \\
\hline
\end{tabular}
\end{table}

We compare the classification accuracy of different methods in the three large-scale datasets, as shown in Table \ref{tab:softmaxpre}. It's very clear that our KNN softmax gets the same accuracy as the full softmax on all datasets. Different from our lossless KNN graph, the Hashing Forest adopted by selective softmax can not make sure that all the true active classes are recalled during training process. MACH performs worse than selective softmax since it uses a divide-and-conquer algorithm to approximate the original classification, which can not guarantee the final accuracy. Our KNN softmax takes advantage of linear KNN graph to ensure no nearest neighbors are lost. The experimental results demonstrate that our KNN strategy makes sense.

\begin{table}[h]\fontsize{9.5pt}{\baselineskip}\selectfont
\centering
\caption{\label{tab:softmaxspeed} The throughput improvement of KNN softmax in the three datasets.}
\begin{tabular}{|l||c|c|c|} \hline
\#methods &{\bf 1M} & {\bf 10M} & {\bf 100M} \\ \hline
Full Softmax   	&  1.0$\times$    & 1.0$\times$ & 1.0$\times$  \\
KNN Softmax    &  1.2$\times$ & 1.5$\times$ & 3.5$\times$ \\
\hline
\end{tabular}
\end{table}

We set the throughputs of full softmax in three datasets as baselines, and evaluate the speed-up of our KNN softmax. Selective softmax and MACH are ignored for their accuracy. Table \ref{tab:softmaxspeed} shows that at the scale of 100M, a speed-up of more than three times is achieved by our approach. Meanwhile, the speed-up tends to be more significant as the size of the dataset is increasing. This is because as the size of the dataset increased, the proportion of the time spent in the softmax stage will also increase, resulting in an increasing speed-up ratio of softmax stage. Thanks to the total GPU pipeline, our KNN softmax is superior to other state-of-the-art methods in consideration of both accuracy and computational efficiency.

\subsection{Evaluation of Communication Strategy}
In this part, we evaluate the large-scale classification training throughput speedup with our proposed communication strategies.

\textbf{Effect of hybrid parallel pipelining. } We compare the hybrid parallel pipelining training throughout with hybrid parallel baseline on the three large-scale datasets. As shown in Table \ref{tab:commspeed}, the overlapping achieves 4.2\%, 4.7\%, 5.4\% performance boost relatively. We tuned the micro-batch size for more network bandwidth usage.

\begin{table}[h]\fontsize{9.5pt}{\baselineskip}\selectfont
\centering
\caption{\label{tab:commspeed} The training speedup with communication optimization in the three datasets.}
\begin{tabular}{|l||c|c|c|} \hline
\#methods &{\bf 1M} & {\bf 10M} & {\bf 100M} \\ \hline
hybrid parallel baseline    &  -    & - & -  \\
+ overlapping    &  1.042$\times$ & 1.047$\times$ & 1.054$\times$ \\
+ layer-wise sparsification &  1.162$\times$ & 1.146$\times$ & 1.123$\times$ \\
\hline
\end{tabular}
\end{table}

\begin{table}[h]\fontsize{9.5pt}{\baselineskip}\selectfont
\centering
\caption{\label{tab:commacc}The training accuracy with layer-wise sparsification in the three datasets.}
\begin{tabular}{|l||c|c|c|} \hline
\#methods &{\bf 1M} & {\bf 10M} & {\bf 100M} \\ \hline
baseline    & 87.43\%  & 81.01\%  & 74.52\%   \\
layer-wise sparsification    & 87.40\%  & 81.05\%  & 74.45\%  \\ 
\hline
\end{tabular}
\end{table}

\begin{table}[h]\fontsize{9.5pt}{\baselineskip}\selectfont
  \centering
  \caption{\label{tab:topk}The wall clock time with different top-$k$ methods (average of 1000 trials). }
  \begin{tabular}{|l||c|c|c|} \hline
  \#methods & time(ms) \\ \hline
  for-loop baseline                   & 204.58     \\
  sampling top-$k$ \cite{lin2017deep} & 83.27     \\
  divide-and-conquer top-$k$          & 36.08     \\
  + tensor grouping                   & 11.81     \\
  \hline
  \end{tabular}
  \end{table}

\textbf{Effect of gradient sparsification. } As shown in Table \ref{tab:commspeed}, combining our efficient hybrid parallel pipeline with gradient sparsification can accelerate training throughput up to 1.123$\times$ in the SKU-100M dataset. Layer-wise top-$k$ sparsification updates partial parameters in a single iteration. We also evaluate the influence on final classification accuracy. Table \ref{tab:commacc} shows introducing layer-wise top-$k$ gradient sparsification in our hybrid parallel framework causes no accuracy degradation in all three datasets. Table \ref{tab:topk} shows the efficiency of our proposed top-$k$ selection method, which can save 94.2\% wall clock time compared with plain for-loop implementation and 7 $\mathbf{\times}$ faster than sampling top-$k$ implementation.

\subsection{Evaluation of Fast Continuous Convergence}
In this subsection, experiments are conducted to show the efficiency of the fast continuous convergence strategy.

% \subsubsection{Baselines}
For the comparison with baselines and the fast continuous convergence strategy, we combine the aforementioned optimization methods with KNN Softmax and communication/computation overlapping to ensure no other factor is incorporated. We evaluate the accuracy and training speed on the following methods: 1) Piecewise decay, the traditional learning rate decay policy, which decays the learning rate by a factor of 1/10 for every five epochs. 2) Adam \cite{kingma2014adam}, of which the initial learning rate is set as 1e-3. 3) FCCS without batch size policy, $B_{max}^1 = 64B_{min}^1 = 64 B_0$, which means only the learning rate policy in FCCS is kept. 4) FCCS, our proposed method, $B_{max}^1 = 64B_{min}^1 = 64B_0$, $T_{final} = 8$. We also keep the same initial batch size $B_0=4096$ and the same initial learning rate $\eta_{0}=0.4$ for each method in each task.

%\subsubsection{Accuracy results}
We compare the accuracy of different convergence strategy on each large classification tasks. As shown in Table \ref{tab:convergence_acc}, the piece-wise decay achieves the best accuracy on all three tasks, and FCCS gets very similar and competitive results, which proves the effectiveness of the adjustment of batch size. Compared with the one without batch size adjustment, FCCS can improve the accuracy from 68.12\% to 87.40\% with the batch size increase policy, which further presents the power of FCCS. The same improvements can be found in the other two large classification tasks too. These results all indicate that the batch size increase policy in FCCS can take the place of the learning rate decay policy in traditional methods. Otherwise, another baseline method, Adams brings obvious loss of accuracy in nearly all the three tasks.

\begin{figure}%[h]
  \centering
  \includegraphics[width=\linewidth]{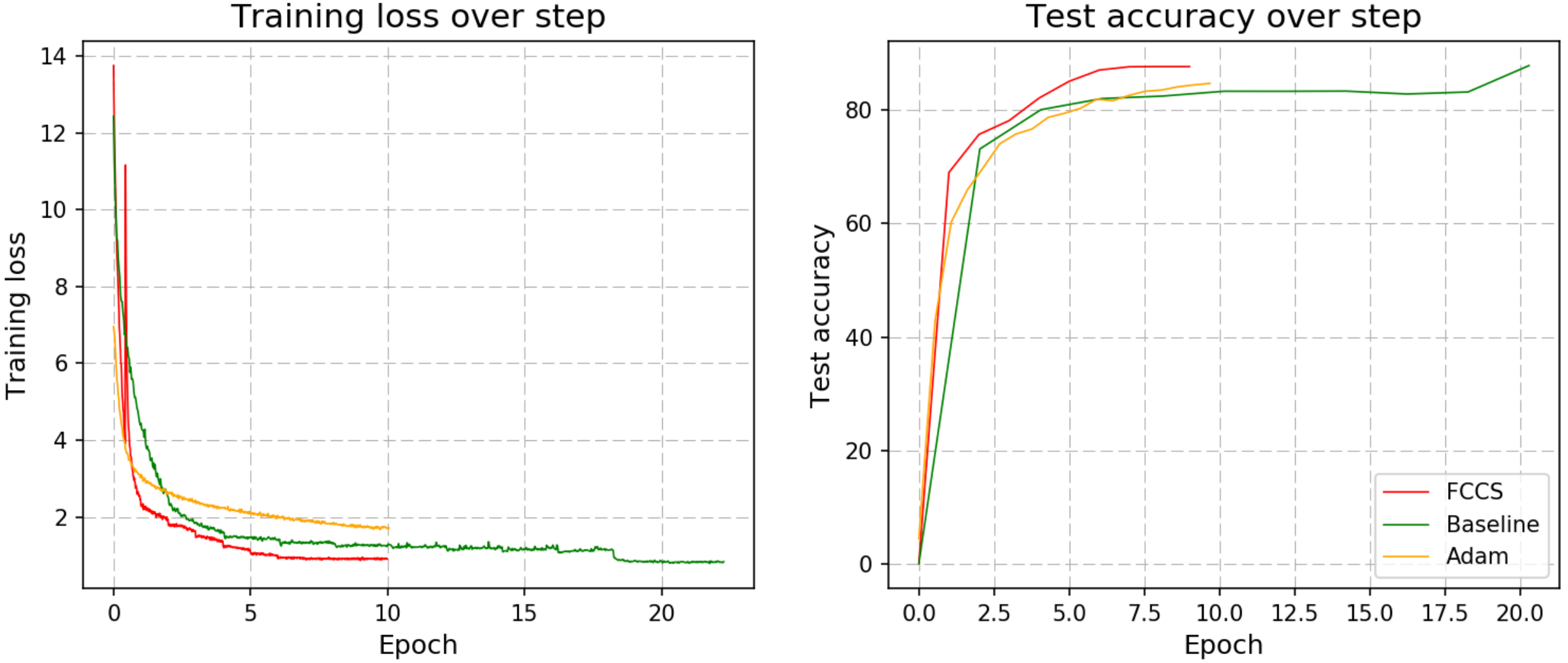}
  \caption{Compare the convergence speed of FCCS and the traditional piece-wise decay learning rate policy.}
  \label{fccs}
\end{figure}

\begin{figure} %[h]
  \centering
  \includegraphics[width=\linewidth]{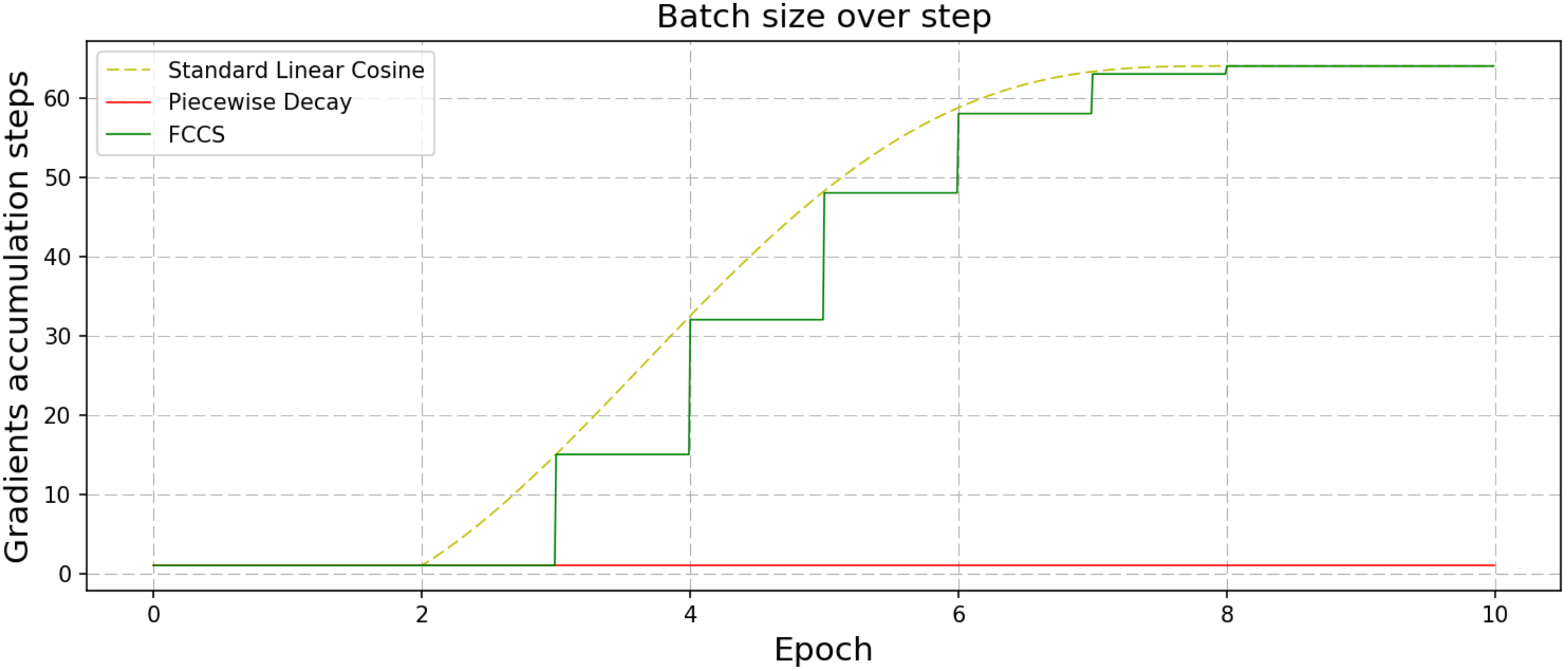}
  \caption{Compare the batch size adjustment of FCCS and the traditional piece-wise decay learning rate policy.}
  \label{FCCS_desc}
\end{figure}

As shown from Figure \ref{fccs}, in the task of 1M classification, the FCCS can reach the final accuracy 87.40\% in 8 epochs while the baseline reaches 87.46\% in nearly 20 epochs. Since the total training procedure speeds up to 2.5 times, 0.4\% loss in test accuracy is tolerable. As shown from Figure \ref{FCCS_desc}, which demonstrates the difference between FCCS and the piece-wise decay on batch size adjustment, we increase the batch size at the beginning of every epoch to simulate the continuous change of linear cosine curve. Note that if we decrease the steps of the plateau in piece-wise decay to make training faster, the final accuracy may become lower and it requires times of effort to adjust the learning rate. Besides, the adaptive optimizer may convergence fast at the begging of training, but finally it brings noticeable accuracy loss compared with our method. The reason that our method outperforms Adam is we change the batch size at a proper range to make it neither too large nor too small.

\begin{table}[h]\fontsize{9.5pt}{\baselineskip}\selectfont
\centering
\caption{\label{tab:convergence_acc} The test accuracy of different training methods in the three datasets.}
\begin{tabular}{|l||c|c|c|} \hline
\#methods                      & 1M      & 10M     & 100M      \\ \hline
FCCS without batch size policy & 68.12\% & 62.77\% & 57.64\%   \\
FCCS                           & 87.40\% & 80.62\% & 74.14\%   \\ \hline
Piecewise decay                & 87.46\% & 80.99\% & 74.51\%   \\
Adam \cite{kingma2014adam}     & 85.16\% & 78.57\% & 72.12\%   \\
\hline
\end{tabular}
\end{table}

\subsection{Evaluation of Extreme Classification System}
As mentioned above, we deploy all the proposed methods together in our extreme classification system to train a classifier of 100 million classes on the SKU-100M dataset. Since these methods are orthogonal in different aspects, we could make full use of them to maximize the training speed of our system.

\begin{figure} %[b]
  \centering
  \includegraphics[width=\linewidth]{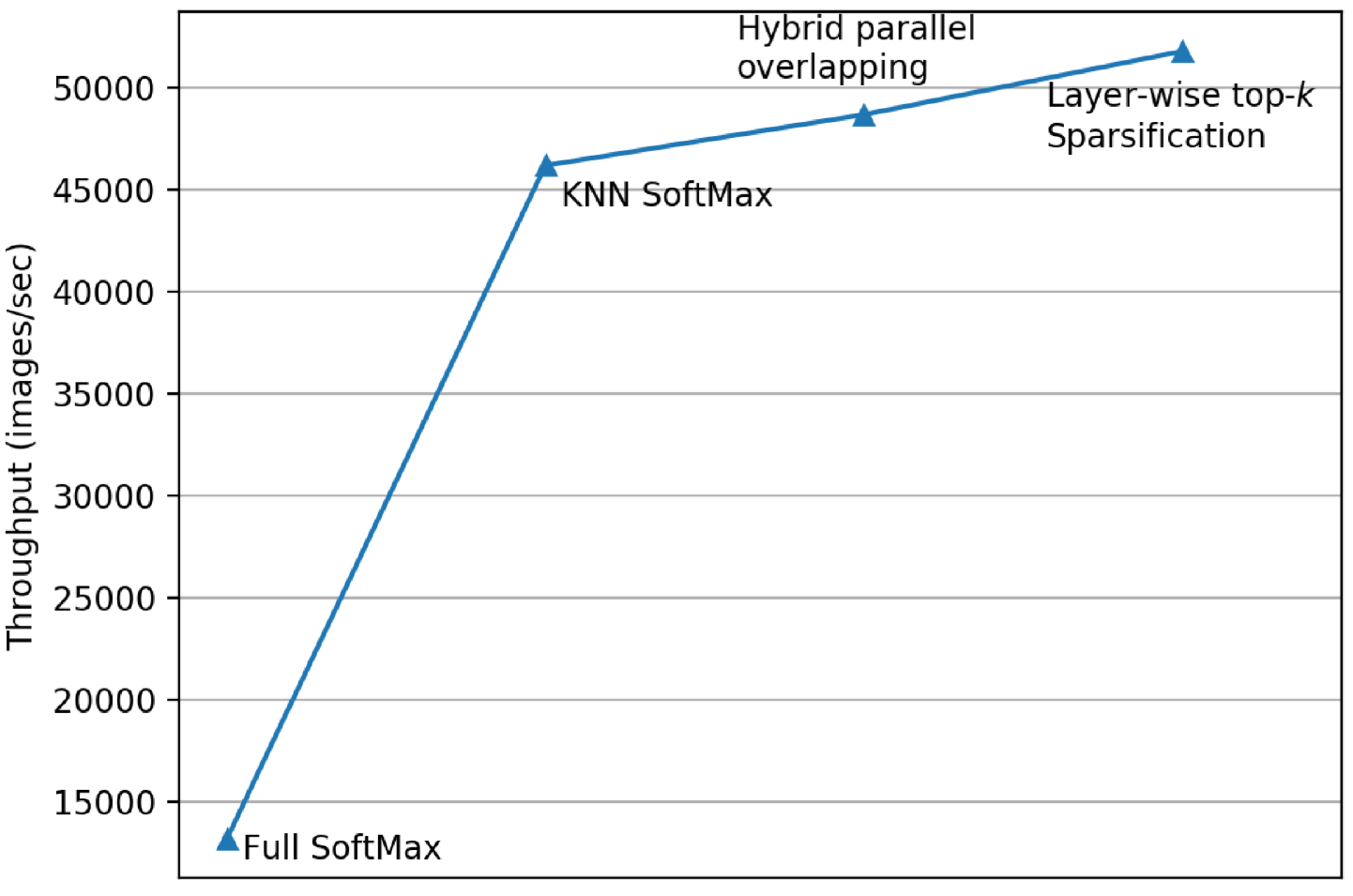}
  \caption{The training speedup with proposed methods.}
  \label{fig:throughput}
\end{figure}

As depicted in Figure \ref{fig:throughput}, we present the system throughput by adding KNN softmax, hybrid parallel overlapping, and layer-wise top-$k$ gradient sparsification sequentially. Compared with the full softmax method as the baseline, the final throughput of our system reaches the improvement of 3.9$\times$ (about 51800 images/sec).

We also adopt the fast continuous convergence strategy (FCCS) to accelerate convergence, which could reduce training iterations from 20 to 8 epochs as 2.5$\times$ speed-up equivalently.

The final results are shown in Table \ref{tab:sysspeed}. Compared with the naive softmax training without FCCS, our proposed method could reduce total training time to five days while reaching a comparable accuracy.

\begin{table}[h]\fontsize{9.5pt}{\baselineskip}\selectfont
\centering
\caption{\label{tab:sysspeed} Final results on SKU-100M dataset.}
\begin{tabular}{|l||c|c|c|} \hline
\#methods         & training time & accuracy \\ \hline
Baseline          & 45 days       & 74.54\%  \\
Proposed Method   & 5 days        & 74.14\%  \\
\hline
\end{tabular}
\end{table}

\subsection{Deployment}

After finishing training the classifier of 100 million classes, we deploy the large model by using the in-house retrieval system \cite{zhao2019large}. For the weight of the fully connected layer $\textbf{W}$, we treat the weight vector $\textbf{w}_{j}$ as the feature embedding for the $j$-th class. Then we use all of the embeddings to build a graph index for classifying images. The online classification process is described as follows: 1) Get the query image and pre-processing it. 2) Feed the query image into the feature extraction model to get the feature embedding. 3) Use the feature embedding to compare and search across the whole index to find the nearest neighbor as the final class. 4) Return the classification result. It only takes one GPU to deploy a feature extraction model with the retrieval system. Moreover, we could add more GPUs incrementally to deal with a large number of queries.

\section{Conclusion}
In this work, we propose an extreme classification system at 100 million class scale. We deploy a KNN softmax implementation to reduce GPU memory consumption and computation costs. As the system is running on the in-house GPU cluster, we design a new communication strategy that contains a hybrid parallel overlapping pipeline and layer-wise top-$k$ gradient sparsification to reduce communication overhead. We also propose a fast continuous convergence strategy to accelerate training by adaptively adjusting learning rate and updating parameters. All of these methods try to improve the speed of training the extreme classifier. The experimental results show that using an in-house 256 GPUs cluster, we reduce the total training time to five days and reach a comparable accuracy with the naive softmax training process.

\bibliographystyle{ACM-Reference-Format}
\bibliography{references}

% \appendix

\end{document}